# Neural network with optimal neuron activation functions based on additive Gaussian process regression

Sergei Manzhos[1], Manabu Ihara[2]

School of Materials and Chemical Technology, Tokyo Institute of Technology, Ookayama 2-12-1, Meguro-ku, Tokyo 152-8552 Japan

## Abstract

Feed-forward neural networks (NN) are a staple machine learning method widely used in many areas of science and technology. While even a single-hidden layer NN is a universal approximator, its expressive power is limited by the use of simple neuron activation functions (such as sigmoid functions) that are typically the same for all neurons. More flexible neuron activation functions would allow using fewer neurons and layers and thereby save computational cost and improve expressive power. We show that additive Gaussian process regression (GPR) can be used to construct optimal neuron activation functions that are individual to each neuron. An approach is also introduced that avoids non-linear fitting of neural network parameters. The resulting method combines the advantage of robustness of a linear regression with the higher expressive power of a NN. We demonstrate the approach by fitting the potential energy surfaces of the water molecule and formaldehyde. Without requiring any non-linear optimization, the additive GPR based approach outperforms a conventional NN in the high accuracy regime, where a conventional NN suffers more from overfitting.

---

[1] E-mail: manzhos.s.aa@m.titech.ac.jp
[2] E-mail: mihara@chemeng.titech.ac.jp

# 1 Introduction and methods summary

Feed-forward neural networks (NN)[1] are a staple machine learning method widely used in many areas of science and technology, including computational chemistry and physics and materials modeling, notably for regression problems.[2–12] A standard all-connected single-hidden layer NN of $x \in R^D$ can be expressed as

$$F_k(x) = c_{out}\sigma_{out}\left(\sum_{n=1}^{N} c_{nk}\sigma_n(w_n x + b_n) + b_{out}\right)$$

(1.1)

where $\sigma_n$ are univariate neuron activation functions, $w_n$ are the weights, and $b_n$ are the biases of the so-called hidden layer. Typically the same $\sigma_n$ is used for all neurons, i.e. $\sigma_n(x) = \sigma(x)$. The output neuron $\sigma_{out}$ (which must be monotonic), the output bias $b_{out}$, and the output wight $c_{out}$ can be omitted (subsumed into the left-hand side) without loss of generality. The subscript $k$ on the left-hand side and coefficients $c_{nk}$ is to indicate that a NN may have multiple outputs; it is omitted in the following without loss of generality of the subsequent discussion. One then can write

$$F(x) = \sum_{n=1}^{N} c_n \sigma_{(n)}(w_n x + b_n)$$

(1.2)

When using $F(x)$ to approximate a smooth function $f(x)$, i.e.

$$f(x) \approx \sum_{n=1}^{N} c_n \sigma_{(n)}(w_n x + b_n)$$

(1.3)

the error will depend on the choices of $\sigma_{(n)}$ and $N$ assuming that optimal weights and biases are used. A single-hidden layer NN of Eq. (1.2) with an $n$-independent *shape* of the neuron activation function is a universal approximator in the sense that for any predefined $\Delta$, there exists a finite $N$ such that $r \equiv |f(\pmb{x}) - \sum_{n=0}^{N} c_n \sigma(\pmb{w_n x} + b_n)| < \Delta$.[13,14] In Eq. (1.3), the notation $\sigma_{(n)}$ is to indicate that the shape of the function may but need not depend on $n$. The expression (1.3) is a linear regression of $f(\pmb{x})$ over a basis $\{\sigma_n\}$ with basis functions

$$\sigma_n = \sigma_{(n)}(\pmb{w_n x} + b_n)$$

(1.4)

The basis functions are $n$-dependent, parameterized via $\pmb{w_n}, b_n$ whether or not the shape of the neuron is $n$-dependent.

Eq. (1.2) is universal approximator for *any* smooth non-linear $\sigma(x)$.[15] This understanding was a conclusion of a series of papers starting from the Kolmogorov theorem,[16] which established the principle of representation of a multivariate function via a superposition of (generally different) univariate functions, followed by works that gradually relaxed the requirements on the neuron activation function.[13,14,17–28] In applications, sigmoidal neuron activation functions[17,29] are widely used, and, less often, simple non-sigmoidal functions that are advantageous for different applications, for example, $\sigma(x) = e^x$ imparts the sum-of-products form to the approximation of $f(\pmb{x})$,[30] which greatly facilitates integration (which is important, for example, for applications in quantum chemistry[31]). The view of a single-hidden layer NN as a basis expansion over a flexible, parameterized basis makes it clear that the expressive power of the method can be maximized by choosing the shapes of $\sigma_n(x)$ which are best in some sense (such as minimizing $r$ for a given $N$) for a regression problem at hand.

One can write Eq. (1.3) as

$$f(\pmb{x}) \approx \sum_{n=1}^{N} c_n \sigma_n(y_n)$$

(1.5)

where $\pmb{y} = \pmb{Wx} + \pmb{b}$, with $\pmb{W}$ a matrix of all $\pmb{w_n}$ and $\pmb{b}$ a vector of all $b_n$. $\pmb{y}$ are in general redundant coordinates. Eq. (1.5) is an *additive model* in $\pmb{y}$. We note that a multilayer NN can also be thought of as an additive model in new coordinates:

$$f(x) = \sum_{k_n=1}^{N_n} w_{k_n}^{(n)} \sigma_{n,k_n}(y_{k_n})$$

$$y_{k_n} = \sum_{k_{n-1}=1}^{N_{n-1}} w_{k_{n-1}}^{(n-1)} \sigma_{n-1,k_{n-1}} \left( \cdots \sum_{k_1=1}^{N_1} w_{k_2}^{(2)} \sigma_{1,k_1} \left( \sum_{i=0}^{d} w_{k_1 i}^{(1)} x_i \right) \right)$$

(1.6)

For brevity of notation, here we subsumed biases into weights (which can be done by introducing dummy variables $x_0$ set to 1) without loss of generality. While the (generally) redundant coordinates $y$ of Eq. (1.6) are non-linear functions of the original coordinates $x$, the $y$ of the single-hidden layer NN, Eq. (1.5), are linear in $x$. In conventional NNs, parameters $W$ and $b$ need to be optimized in a non-linear optimization, which dominates the cost of building a NN model and can cause overfitting. One reason while the same simple-form activation functions are typically used for all neurons is the possibility to simplify this optimization with backpropagation.[32,33] NNs which do not fit the nonlinear parameters at all have been proposed – effectively using random basis functions that are non-optimal.[34]

An additive model is a particular case of the high-dimensional model representation (HDMR)[35–38]

$$f(x) \approx f_0 + \sum_{i=1}^{D} f_i(x_i) + \sum_{1 \leq i < j \leq D} f_{ij}(x_i, x_j) + \cdots + \sum_{\{i_1 i_2 \ldots i_d\} \in \{12 \ldots D\}} f_{i_1 i_2 \ldots i_d}(x_{i_1}, x_{i_2}, \ldots, x_{i_d})$$

(1.7)

This is an expansion over orders of coupling of the variables $x_i$. Including the term with $d = D$, this model is exact; when maximum considered $d < D$, it is approximate. When $d = 1$, one obtains a plain, first-order additive model. The advantage of using HDMR is that for most applications, the importance of terms drops rapidly with $d$, and a maximum $d$ of anywhere between 2 and 4 is often sufficient.[35,39–43] Then one works only with low-dimensional component functions, which is advantageous for both building and using the model.

Moreover, when the data used to train the model are sparse, the data density puts a limit on the highest recoverable order of coupling,[39,41] so that an HDMR with a low $d$ may be as or more accurate than a full-dimensional model. When $d$ is sufficiently low, this approximation can avoid overfitting even with few data, as low-order terms are well-defined with few data.[35,40,41,44] In this sense an additive model with $d = 1$ is the most robust, but while a $d$ value of 2-4 is often sufficient, the accuracy of the first-order additive model *in the original coordinates* is often not good enough for use in applications, see Refs. [40,41,45] for some examples. A disadvantage of HDMR is a combinatorial scaling of the number of terms with both $D$ and $d$, although some terms can be neglected.[40] This disadvantage is nullified of one stays at $d = 1$. A first order additive model in redundant coordinates, however, even if those coordinates linearly depend on the original coordinates, is a universal approximator as it has the form of a single-hidden layer NN as per Eqs. (1.3-1.5). We note that HDMR in general allows constructing low-dimensional terms that maximize the accuracy of the approximation (1.7) in all space, or on all training datapoints however distributed in space; the terms of HDMR are in general *not* slices of $f(x)$ along sub-dimensional hypersurfaces[35,36,38] (contrary, for example, to the terms of an N-mode appriximation[46]).

Building an optimal additive model means finding optimal $y_n$ and $\sigma_n(y_n)$, which are also the component functions $f_i(x_i)$ of a 1-$d$ HDMR (we will not explicitly consider the constant term as it is trivial and can be subsumed into the left hand side or into $f_i(x_i)$). We previously proposed HDMR with machine-learned (ML) component functions, specifically, HDMR-NN[39,42] and HDMR-GPR[40,41,45] combinations in which HDMR component functions are represented with, respectively, neural networks and Gaussian process regressions[47] (GPR). ML component functions are optimal in the least-squares sense. In particular, HDMR-GPR (or additive GPR[48]) offers an easy way to build the component functions due to the non-parametric nature of GPR, whereby one defines an HDMR-type kernel and then the HDMR model is obtained by linear algebra. The GPR approximation has the form

$$f(\boldsymbol{x}) = \sum_{n=1}^{M} b_n(\boldsymbol{x}) c_n = \sum_{n=1}^{M} k(\boldsymbol{x}, \boldsymbol{x}^{(n)}) c_n = \boldsymbol{Bc}$$

(1.8)

where $k(x, x')$ is the kernel (covariance function between points $x$ and $x$') and the coefficients $c$ are obtained as

$$c = K^{-1}f$$

(1.9)

where

$$K = \begin{pmatrix} k(x^{(1)}, x^{(1)}) + \delta & k(x^{(1)}, x^{(2)}) & \cdots & k(x^{(1)}, x^{(M)}) \\ k(x^{(2)}, x^{(1)}) & k(x^{(2)}, x^{(2)}) + \delta & \cdots & k(x^{(2)}, x^{(M)}) \\ \vdots & & \ddots & \vdots \\ k(x^{(M)}, x^{(1)}) & k(x^{(M)}, x^{(2)}) & \cdots & k(x^{(M)}, x^{(M)}) + \delta \end{pmatrix}$$

(1.10)

is the covariance matrix between $M$ training points - known samples $f_n = f(x^{(n)})$. The regularization parameter $\delta$ is typically added on the diagonal to improve stability and generalization. $f$ is the vector of all $f_n$. The covariance function is usually chosen as one of the Matern family of functions,[49]

$$k(x, x') = A \frac{2^{1-\nu}}{\Gamma(\nu)} \left( \sqrt{2\nu} \frac{|x - x'|}{l} \right)^\nu K_\nu \left( \sqrt{2\nu} \frac{|x - x'|}{l} \right)$$

(1.11)

where $\Gamma$ is the gamma function, and $K_\nu$ is the modified Bessel function of the second kind. At different values of $\nu$, this function becomes a squared exponential ($\nu \to \infty$), a simple exponential ($\nu=1/2$) and various other widely used kernels (such as Matern3/2 and Matern5/2 for $\nu = 3/2$ and $5/2$, respectively). It is typically preset by the choice of the kernel, and the length scale $l$ and prefactor $A$ (if needed) are hyperparameters. By choosing

$$k(x, x') = \sum_{\{i_1 i_2 \ldots i_d\} \in \{12 \ldots D\}} k_{i_1 i_2 \ldots i_d}(x_{i_1 i_2 \ldots i_d}, x'_{i_1 i_2 \ldots i_d})$$

(1.12)

where $\boldsymbol{x}_{i_1 i_2 \ldots i_d} = (x_{i_1}, x_{i_2}, \ldots, x_{i_d})$ and $k_{i_1 i_2 \ldots i_d}$ is one of Matern kernels in $d$ dimensions, one obtains an HDMR-type representation of $f(\boldsymbol{x})$:[45,48]

$$f(\boldsymbol{x}) = \sum_{\{i_1 i_2 \ldots i_d\} \in \{12 \ldots D\}} f_{i_1 i_2 \ldots i_d}(\boldsymbol{x}_{i_1 i_2 \ldots i_d})$$

(1.13)

where the component functions $f_{i_1 i_2 \ldots i_d}(\boldsymbol{x}_{i_1 i_2 \ldots i_d})$ are least squares solutions and are in this sense optimal.

For given $\boldsymbol{y}$, i.e. for given $\boldsymbol{W}, \boldsymbol{b}$, and $N$, HDMR-GPR thus offers a natural way to construct $\sigma_n(x)$ i.e. to obtain a single-hidden layer NN with optimal activation functions of individual neurons. In this paper, we show that this is indeed a viable strategy and that thus constructed NN with 1-$d$ HDMR based neurons can achieve or surpass the quality of representation achievable with traditional ML techniques. We also avoid non-linear optimization of $\boldsymbol{W}$ and $\boldsymbol{b}$ and instead propose a rule-based approach to define $\boldsymbol{y}$. The method thus involves only linear operations while obtaining a non-linear model. The absence of non-linear optimization of many $\boldsymbol{W}$ and $\boldsymbol{b}$ removes an important argument for the use of the same, simple $\sigma(x)$ for all neurons. We first demonstrate the approach of using HDMR-GPR-derived neurons by fitting the potential energy surface (PES) of the water molecule in the spectroscopically relevant region of the configuration space. This example is chosen because (i) in spite of the relatively low dimensionality, it is sufficient for the present purpose of demonstration of the idea; (ii) a high accuracy (on the order of 1/10,000[th] of the range) is required and is known to be obtainable with ML techniques like NN or GPR, and there is good intuition what constitutes a good or bad accuracy; (iii) additive (HDMR) models of different orders $d$ were previously reported[41] and can be compared to; (iv) we chose a three-dimensional PES where space sampling density can be high enough to avoid conflation of the issue of the quality of representation with the functional form that we obtain and the issue of data sparsity (which is largely avoided here). We then show that the method works just as well in 6D, on the example of fitting the PES of formaldehyde.

## 2 Details of the methods

All calculations are performed in Matlab with a home-made code. The GPR is performed using Matlab's *fitrgp* function with a custom additive kernel

$$k(\pmb{x}, \pmb{x}') = \sum_{i=1}^{D} k_i(x_i, x_i')$$

(2.1)

where we used square exponential kernels for the components $k_i(x_i, x_i')$,

$$k_i(x_i, x_i') = exp\left(-\frac{(x_i - x_i')^2}{2\exp(l)^2}\right)$$

(2.2)

The features were normalized to unit variance or scaled to unit cube before regression, we therefore used a single value of $l$ for all $i$. The regularization parameter $\delta$ was approximately manually optimized and is set to $1\times10^{-6}$. The results below are presented for $l = 1.0$ for water (where we normalized the inputs) and $l = 0.0$ for formaldehyde (where the inputs were scaled to unit cube). While there is some (minor) variability of optimal $l$ depending on $N$, it is not important for the purpose of this study. The component functions $f_i(x_i)$ of the additive model $f(\pmb{x}) \approx \sum_{i=1}^{D} f_i(x_i)$ are computed as

$$f_i(x_i) = \pmb{K}_i^* \pmb{c}$$

(2.3)

where $\pmb{K}_i^*$ is a row vector with elements $k_i\left(x_i, x_i^{(n)}\right)$, $n$ indexing training points. The vector $\pmb{c}$ is computed as in Eq. (1.9). The variance of the component functions is computed on the training point set. The component functions are linear combinations of smooth kernel functions; they therefore satisfy the conditions for universal approximator when considered as neuron activation functions.[15]

The redundant variables $y$ are constructed by two type of rules, named below "pairwise scheme" and "pseudorandom scheme". In the pairwise scheme, new vectors are defined as averages of all pairs of coordinates and adding them to the coordinate set. The same procedure is then repeated on the new coordinate set in as many cycles as necessary, generating datasets with different $N$. Each cycle thus increments the order of coupling: the first step samples only second order coupling terms (two-dimensional hyperplanes), the second step second and third order (three-dimensional hyperplanes) etc. This scheme works well in low-dimensional cases (such as the 3D example below) but generates a rapidly, stepwise growing number of new coordinates (which can, however, be reduced by using subsets of pairs). The pseudorandom scheme allows defining any number of redundant coordinates. The new coordinates are obtained as $y_n = x^T s_n$ where $s_n$ is an element of a $D$-dimensional pseudorandom sequence. In this scheme, all $y_n$ couple all $x_n$. We use a Sobol sequence.[50] A pseudorandom sequence is used because a random sequence results in an uneven distribution unless the number of elements is large. These procedures generate $W$ and ignore $b$. The biases are unnecessary, as addition or change of any bias value can automatically be compensated for by the change of $\sigma(x)$ by HDMR-GPR. We do not optimize $W$ but show that moderate values of $N$ allow high quality of regression without such optimization. We also show below that $N$ can be reduced by discarding component functions with the lowest magnitudes, based on their variance.

We fitted the potential energy surfaces of $H_2O$ and $H_2CO$. For the description of the datasets and information about the applications of these functions, specifically in computational spectroscopy,[51] see Ref. [52] for $H_2O$ and Ref. [53] for $H_2CO$. We have a total of 10,000 data points in three dimensions representing Radau coordinates of $H_2O$. The Radau coordinates are convenient for quantum dynamics, they are different from the bond coordinates but can mnemonically be thought of as representing the two HO bond lengths and the HOH angle.[54] The values of potential energy range 0-20,000 cm$^{-1}$ and are sampled from the analytic PES of Ref. [55]. The data are available in the supplementary material of Ref.

[40]. For formaldehyde, we have 120,000 data points in six dimensions (representing molecular bonds and angles) with values of potential energy ranging 0-17,000 cm$^{-1}$ sampled from the analytic PES of Ref. [56]. These data are also available in the supplementary material of Ref. [40].

The data points sample the PES around the equilibrium geometry but over a sufficiently wide range to include anharmonic regions and allow calculations of hundreds of vibrational levels.[52] We use a randomly selected subset of $M$ = 500, 1000, or 2000 points for H$_2$O and $M$ = 1000 for H$_2$CO for training and 10,000 points (for H$_2$CO) or the remaining 10,000-$M$ points (for H$_2$O) for testing. The test set is thus much larger than the training set and is sufficiently dense in three dimensions to reliably represent a global quality of the regression. Due to the random nature of the choice of the training points from the full dataset, there is a slight variation of the error from run to run which is insignificant in the context of this study.

## 3 Results

### 3.1 Neural network-like additive Gaussian process regression: testing in 3D

We first perform HMDR-GPR fits to establish the accuracy achievable with additive models and a full-dimensional GPR model. Table 1 lists train and test set errors achievable with different orders of HDMR $d$ and different numbers of training points $M$. Not surprisingly, as the density of sampling increases, the spread between train and test errors decreases. The error generally decreases with $d$. A first-order additive model cannot achieve an error better than 400 cm$^{-1}$ with any $M$ – it is limited by the order of coupling rather than $M$. The second order model's test set error is also limited by the neglected third order coupling and cannot be lower than about 200 cm$^{-1}$. The accuracy of the full-dimensional model is limited by $M$. With $M$ = 1000, a sub-cm$^{-1}$ global accuracy is obtained with the full-dimensional GPR, which is sufficient for the calculation of highly accurate vibrational spectra on the PES.[52] These results can be compared with those obtained in Ref. [40] on the same data with a somewhat different HDMR-GPR scheme using separate GPR instances for component functions rather than an HDMR-type kernel, they are given in the table in parentheses. That works also used scaling of the features to unit cube rather than normalization and different hyperparameters, so numeric differences are expected with the same trend.

Table 1. Train/test set error, in cm$^{-1}$, achieved with different orders $d$ of HDMR-GPR and different numbers of training points $M$. The numbers in parentheses are test set rmse values from Ref. [40] that used a different approach to HDMR-GPR.

| $M$ | $d=1$ | $d=2$ | $d=3$ |
|---|---|---|---|
| 500 | 412 / 499 (460) | 125 / 707 (268) | 0.11 / 1.27 (1.9) |
| 1000 | 438 / 511 (463) | 184 / 361 (259) | 0.17 / 0.53 |
| 2000 | 435 / 455 | 187 / 232 | 0.18 / 0.27 |

We therefore use $M = 1000$ in the following. In this example, we first use the pairwise scheme to define redundant coordinates. Figure 1 shows correlation plots between the reference PES samples and the fits as well as relative importance of the component functions. All correlation plots show errors increasing with energy; this is a consequence of the deliberate distribution of the data that is more dense in the low-energy regions, see Ref. [52] for more details. The presence of a number of points with large errors in the high energy, low data density regions pull up the rmse values for $d = 2$, while improvement at lower energies vs. $d = 1$ is clear. For the case of $d = 1$, plots of component functions are shown in Figure 2.

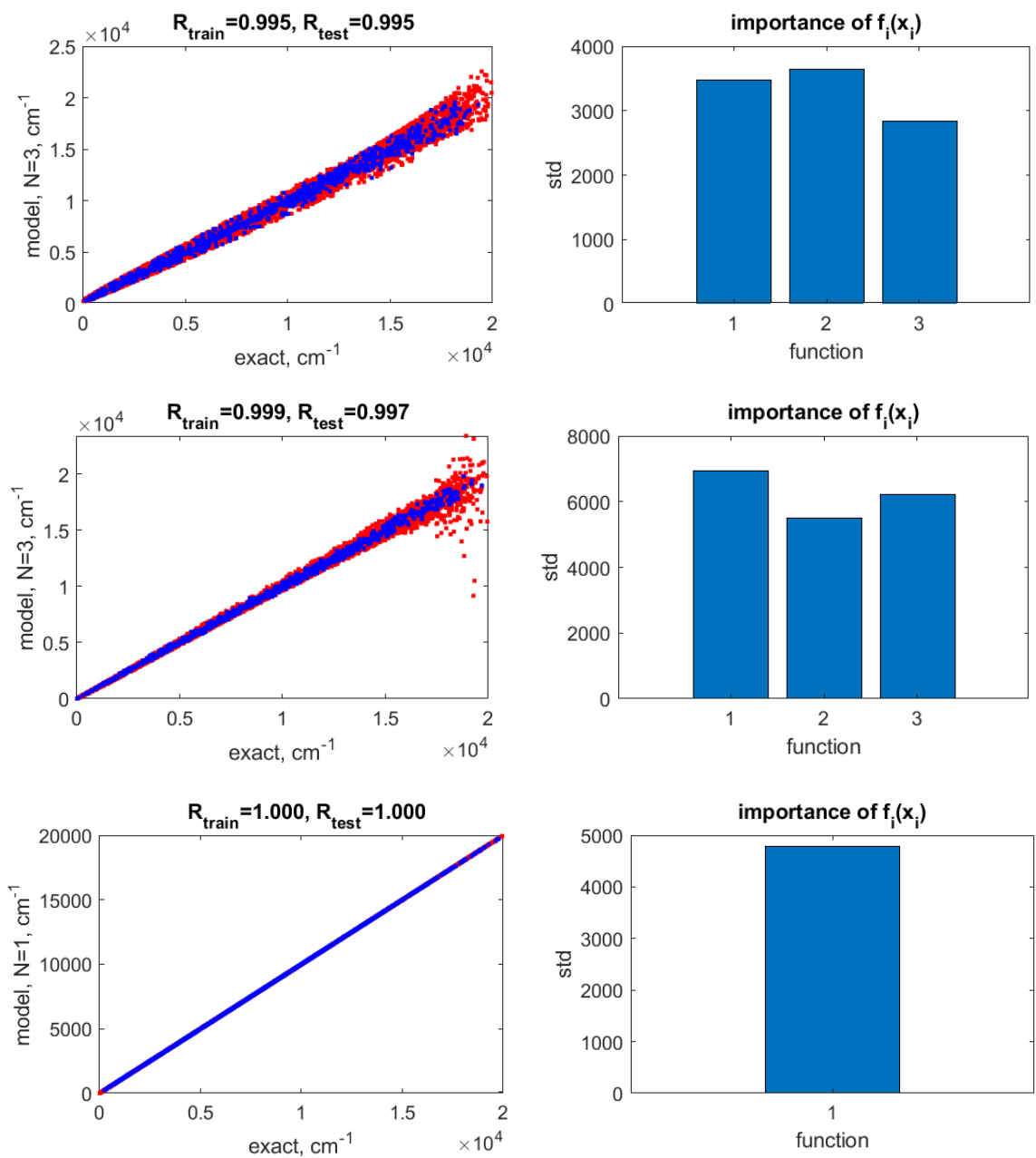

Figure 1. Left: correlation plots between the reference H$_2$O PES samples and the fits, for $M = 1000$ training points and different orders of HDMR $d$. The blue dots are the training points, and the red dots are the testing points. The corresponding correlation coefficients for training and test points are also given. Right: the relative importance of component functions, by the magnitude of their standard deviations ("std"). Top to bottom: $d = 1$, $d = 2$, $d = 3$.

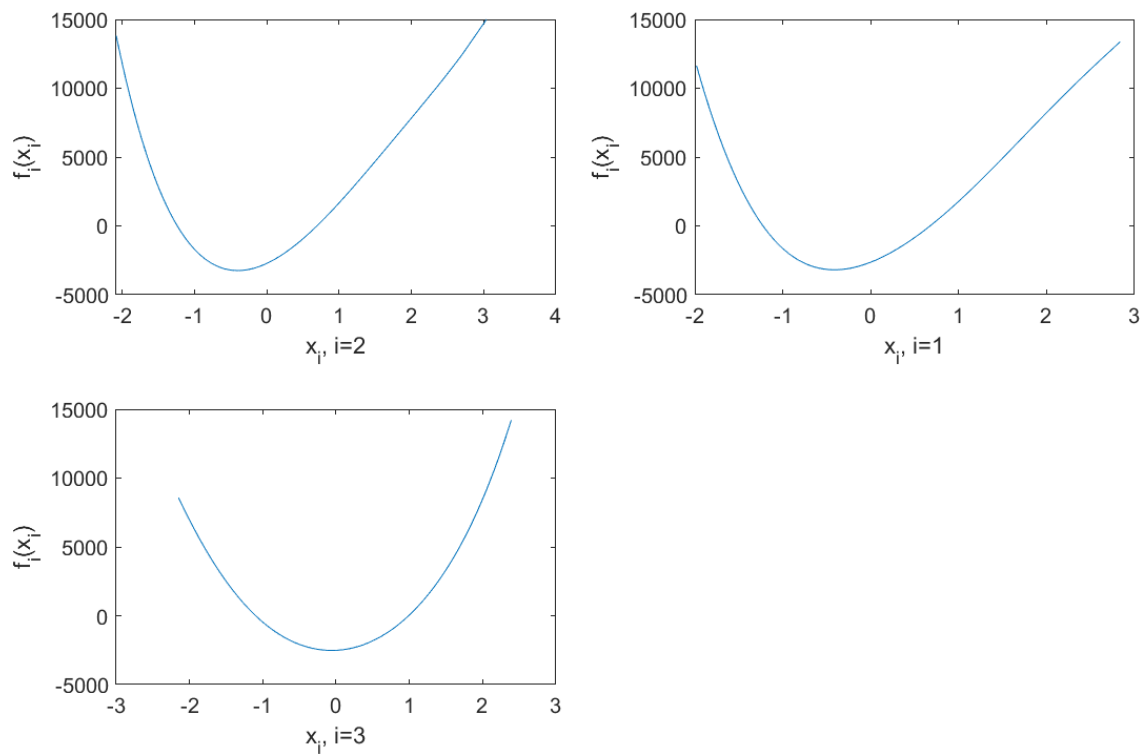

Figure 2. Component functions of the first-order HDMR model, in the order of importance, for the case of $M = 1000$.

We now perform regressions, using $M = 1000$ training data, with the model of Eq. (1.5) for different $N$ that result from applying the pairwise procedure of generating components of $y$ one, two, and three times consecutively. This results in $N = 6, 21$, and 231. While the set of 6 components of $y$ samples second-order coupling only, the sets of 21 and 231 $y_n$ sample third-order coupling terms. The corresponding train / test errors are 219 / 257, 14.8 / 26.6, and 0.35 / 0.79 cm$^{-1}$, respectively. The correlation plots, plots of relative importance of the component functions, and plots of the four most important component functions are shown in Figure 3, Figure 4, and Figure 5. With six terms, the redundant coordinate first-order HDMR model of Eq. (1.5) outperformed a second-order HDMR-GPR; with 21 terms, the test error approaches values suitable for use in applications including spectroscopic applications,[57] and with 231 terms, the model is as good as a high-quality full-dimensional model, either GPR or NN.[52]

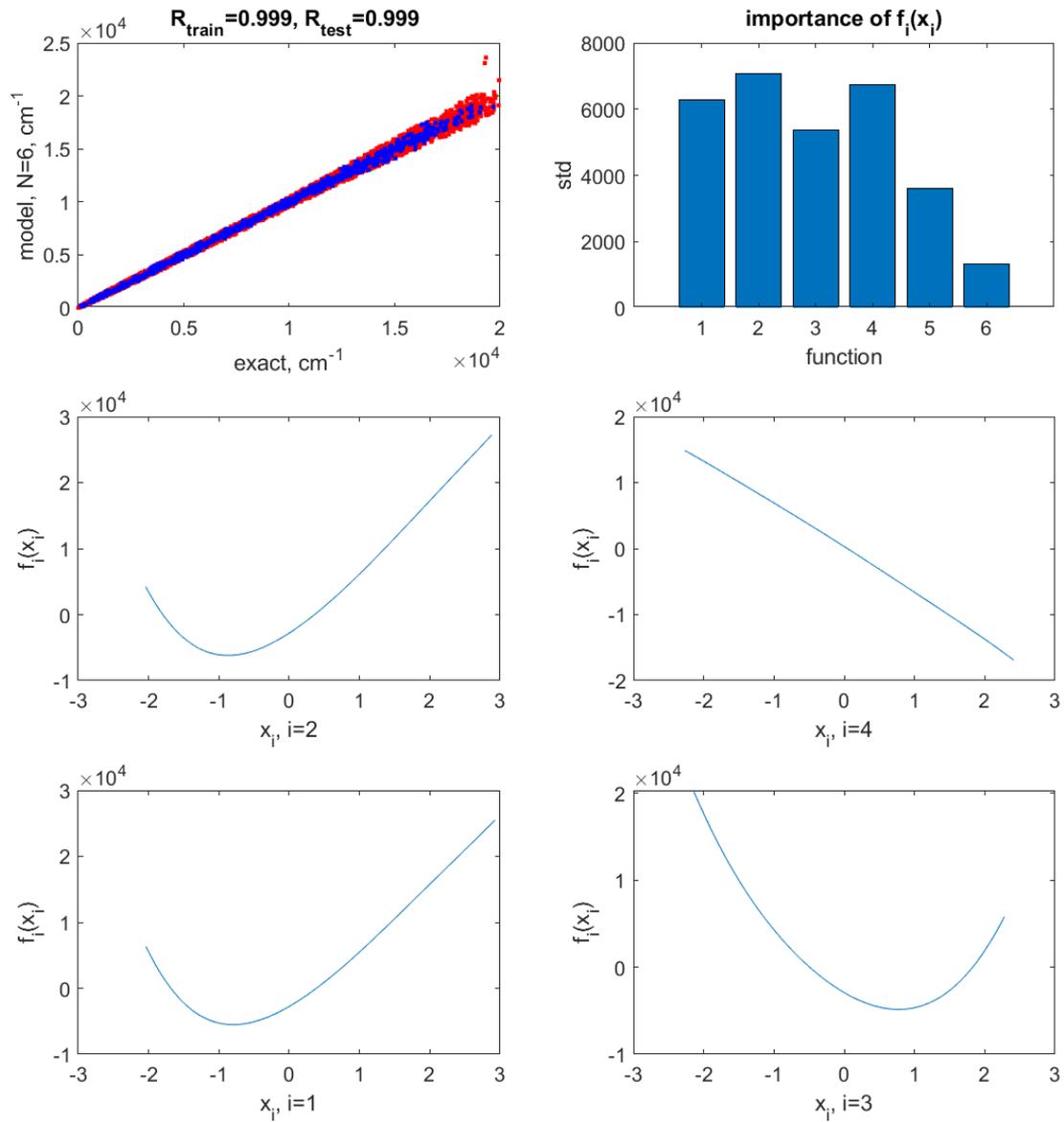

Figure 3. The correlation plot (top left), plot of relative importance of the component functions (top right), and plots of the four most important component functions (by the magnitude of standard deviation, "std") when using a redundant coordinate first-order HDMR model of Eq. (1.5) with $N = 6$ terms. In the correlation plot, the blue dots are the training points, and the red dots are the testing points. The corresponding correlation coefficients for training and test points are also given.

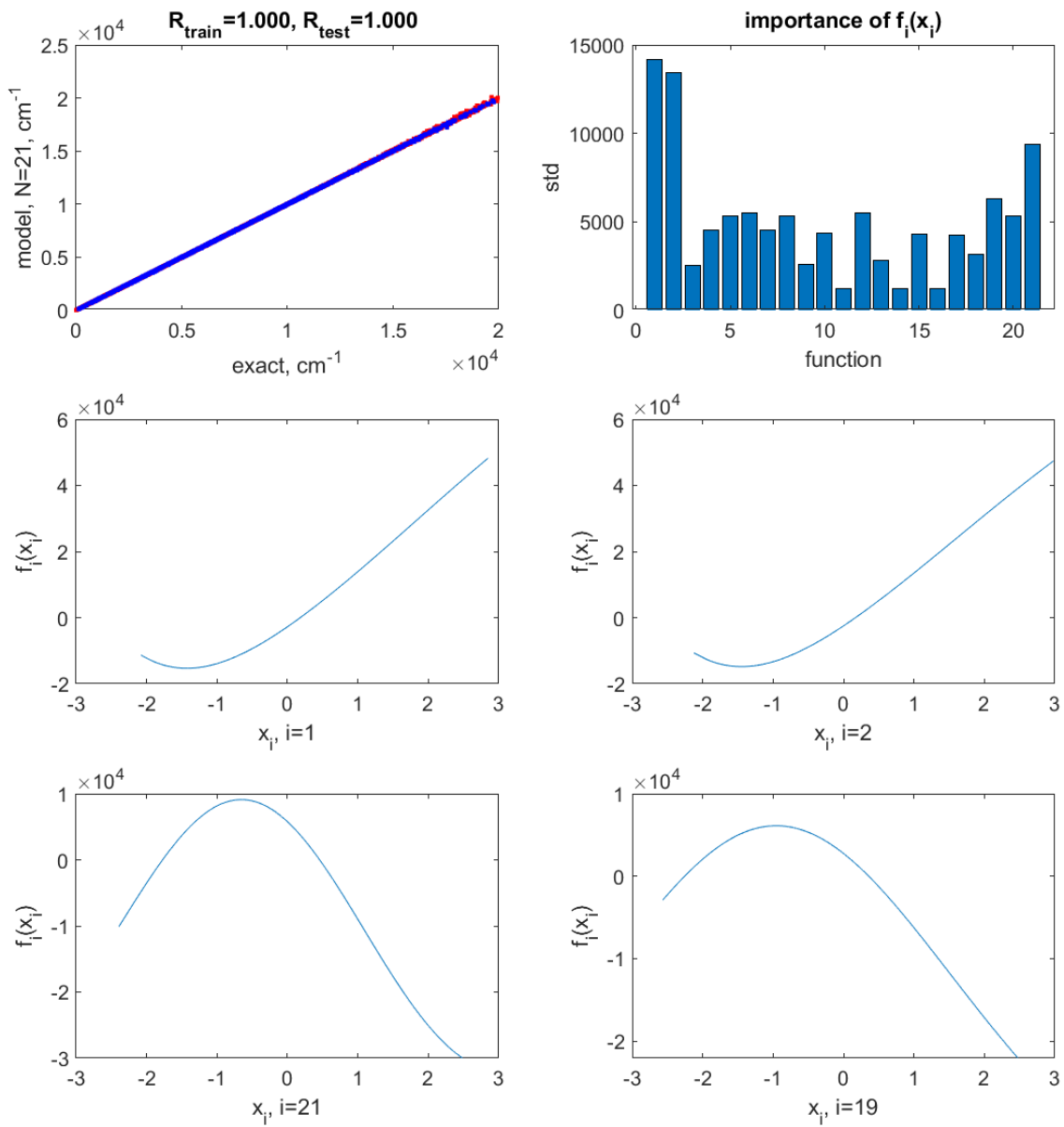

Figure 4. The correlation plot (top left), plot of relative importance of the component functions (top right), and plots of the four most important component functions (by the magnitude of standard deviation, "std") when using a redundant coordinate first-order HDMR model of Eq. (1.5) with $N = 21$ terms. In the correlation plot, the blue dots are the training points, and the red dots are the testing points. The corresponding correlation coefficients for training and test points are also given.

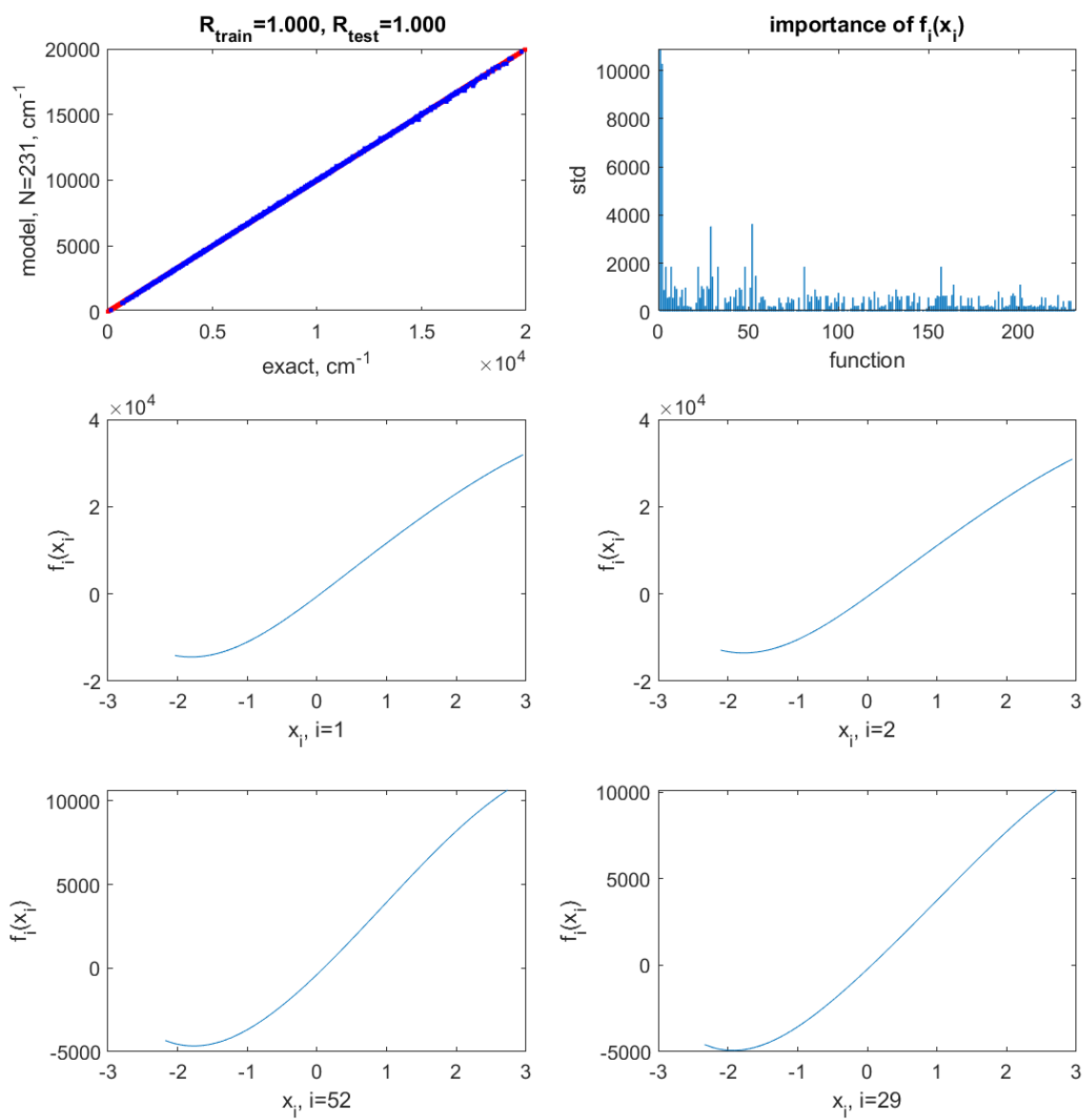

Figure 5. The correlation plot (top left), plot of relative importance of the component functions (top right), and plots of the four most important component functions (by the magnitude of standard deviation, "std") when using a redundant coordinate first-order HDMR model of Eq. (1.5) with $N = 231$ terms. In the correlation plot, the blue dots are the training points, and the red dots are the testing points. The corresponding correlation coefficients for training and test points are also given.

*3.2 Component function pruning*

As the number of terms increases, there appear to be many component functions with small magnitudes. We therefore study the possibility of discarding some of them based on their variance. Figure 6 shows the behavior of the rmse for the training and test sets depending on the number of retained terms out of the 231 terms generated by the pairwise scheme. The error decreases with *N* relatively monotonically until it plateaus at about 0.7 cm$^{-1}$ for the train data and 1.4 cm$^{-1}$ for the test data for *N* between 56 and 86 terms. With 91 terms a second plateau is reached at about 0.35 and 0.8 cm$^{-1}$ for train and test points, respectively. There is no advantage of going beyond 91 terms.

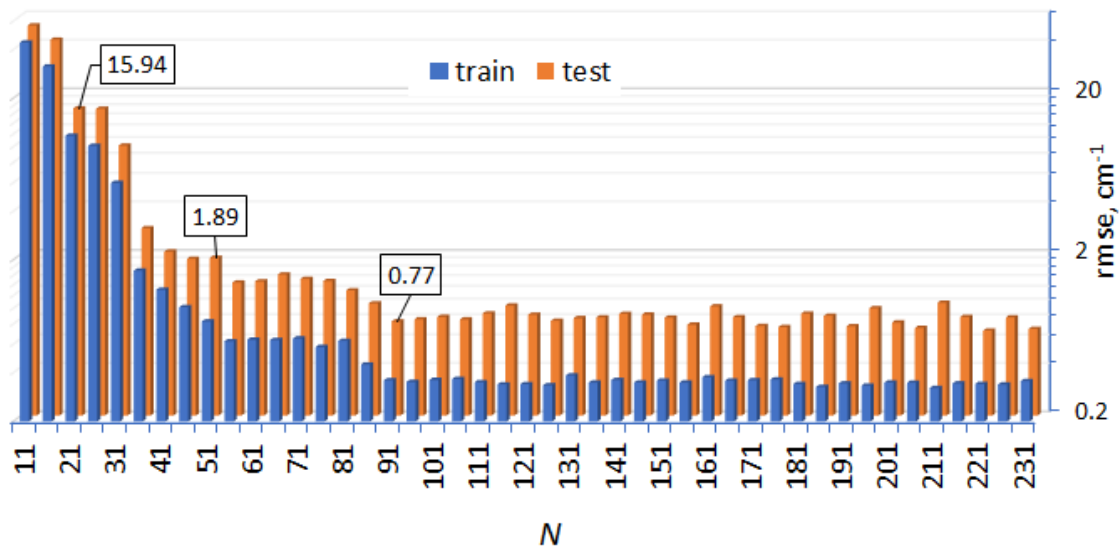

Figure 6. Train and test set errors of the H$_2$O PES as a function of the number of terms *N* using the pairwise scheme of generating neuron arguments. Note the logarithmic scale. The test set rmse values compared in the text with those from a conventional neural network with the same number of neurons are indicated.

For comparison, Figure 7 shows the same plot for the case of using the pseudorandom scheme to generate *y*. To facilitate comparison, we also prune starting from 231 terms. With about 50 terms, a test set error on the order of 0.5 cm$^{-1}$ (and training set error about 0.2 cm$^{-1}$) is obtained with no improvement with more terms. The test set error drops below 2 cm$^{-1}$ after 31 terms. Overall, the pseudorandom scheme performs somewhat better than the pairwise scheme.

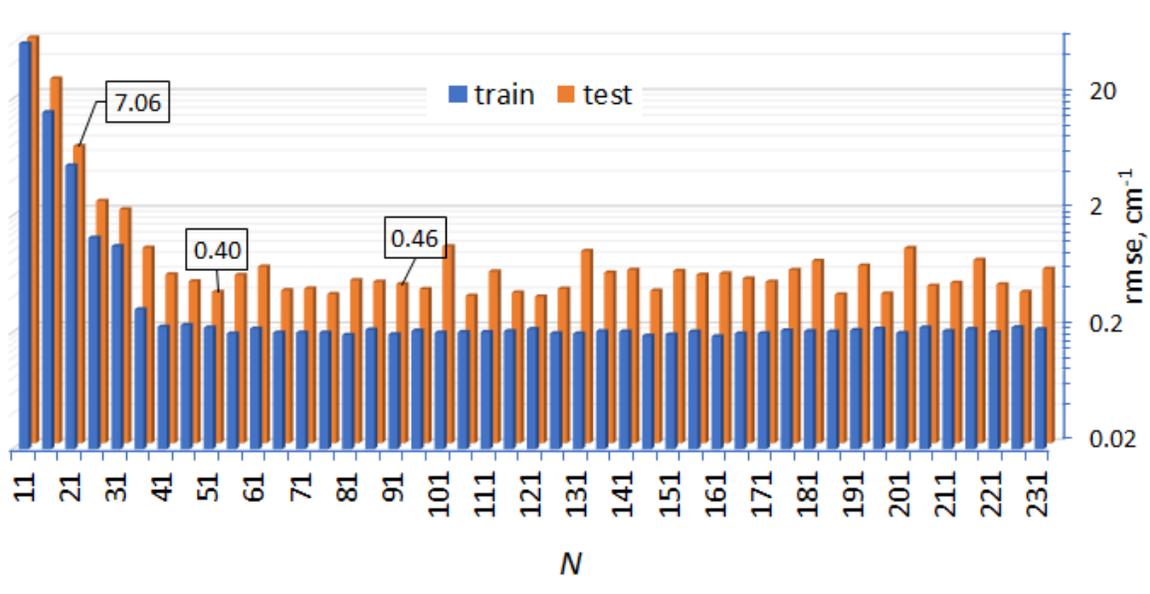

Figure 7. Train and test set errors of the $H_2O$ PES as a function of the number of terms $N$ using the pseudorandom scheme of generating neuron arguments. Note the logarithmic scale. The test set rmse values compared in the text with those from a conventional neural network with the same number of neurons are indicated.

To put these results in perspective, we can compare them to the train and test errors achieved with a conventional single-hidden later NN with sigmoid neurons and with high-quality non-linear optimization of parameters with the Levenberg-Marquardt algorithm,[58] for 1000 cycles (which is sufficient for convergence). Non-linear parameter optimization introduces the risk of local minima. There is a distribution of errors run to run which is significant and far greater than the practically insignificant distribution due to the random selection of the training points. The results are shown in Figure 8 for a statistic over 10 runs. We use 21, 51, and 91 neurons as the numbers of terms around the beginning of the plateaus in Figure 6. With 21 terms / neurons, the mean test error is about 4 cm$^{-1}$, which is better than the test error of about 16 and 7 cm$^{-1}$ obtained with rule-based *y* using the pairwise and pseudorandom scheme of generating neuron arguments, respectively, but the spread run to run in the conventional NN is much more significant, reaching a factor of 6. The spread of values within the plateaus in Figure 6 and Figure 7, which is due to different choices of *y* as well as different draws of the training points, on the other hand, is within a factor of 1.5 and 2, respectively, for all cases.

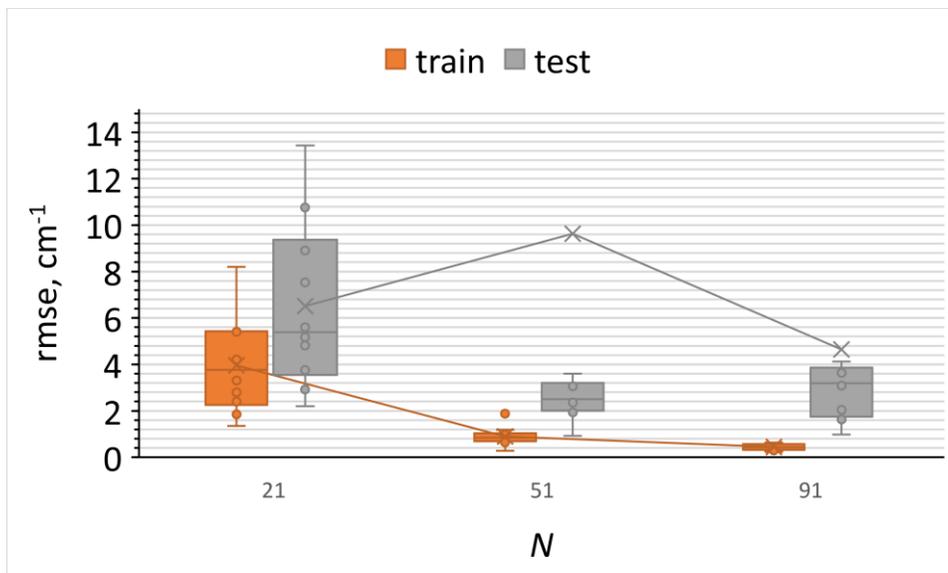

Figure 8. Distributions of train and test set errors from 10 fits of the $H_2O$ PES with a conventional neural network with different numbers of neurons *N*. The cross shows the average error, and the middle line the median error over 10 runs. First and third quartiles are indicated by the box. For 51 and 91 neurons, outlier points at 74.3 and 22.2 cm$^{-1}$, respectively, are not shown on the scale of the plot but affect the average.

Already with 51 neurons, the conventional NN shows a significant overfitting. The mean test error is on the order of 10 cm$^{-1}$ (vs. the average training set error on the order of 1 cm$^{-1}$), due to an outlier (not shown on this scale) at 74.3 cm$^{-1}$. The median error that does not suffer from the outlier at about 2.5 cm$^{-1}$ is still *worse* than with the HDMR-GPR based approach (about 1.9 and 0.4 cm$^{-1}$, with the pairwise and pseudorandom scheme, respectively, and no significant spread run to run). With 91 neurons, where the HDMR-GPR based approach achieves a training set error of 0.36 and 0.19 cm$^{-1}$ and a test set error of 0.77 and 0.46 cm$^{-1}$ (with the pairwise and pseudorandom scheme, respectively), the conventional NN obtains an average (median) error of 0.44 (0.43) for the train set and 4.65 (3.18) cm$^{-1}$ for the test set. The train set error can always be made as small as needed by using more neurons;

the relevant error is the test error, for which there is no improvement vs the NN with 51 neurons. Without requiring any non-linear optimization, the NN with fixed, rule-based *y* (neuron inputs) and HDMR-GPR based neuron activation functions thus outperforms a conventional NN in the high accuracy regime, where a conventional NN suffers more from overfitting. The increase of the number of terms does not lead to any significant worsening of the test set error.

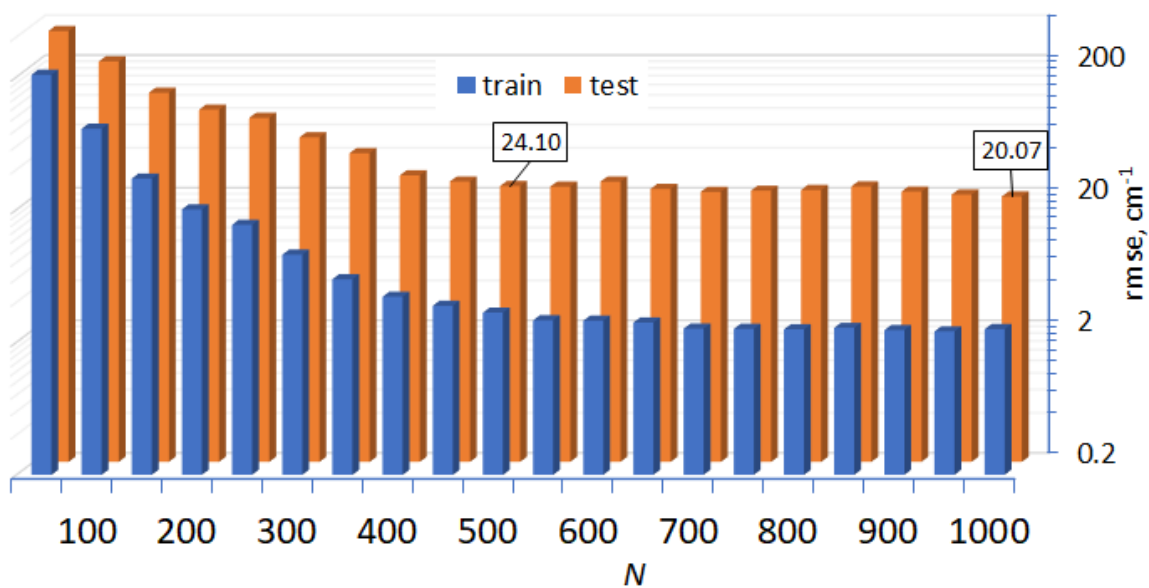

Figure 9. Train and test set errors of the $H_2CO$ PES as a function of the number of terms *N* using the pseudorandom scheme of generating neuron arguments. Note the logarithmic scale.

## 3.3 A 6D example

To demonstrate the performance of the approach in a higher dimensionality, we fit a six-dimensional PES of formaldehyde. In this example, we use the pseudorandom scheme to define the arguments *y* of neuron activation functions, as it has shown a superior performance in the 3D example and allows easier application in higher dimensionality. The results, train and test rmse, are shown as a function of the number of terms / neurons in Figure 9. To put these results in perspective, the full-dimensional GPR using the same number of training points ($M = 1000$) and an optimal length parameter resulted in a test set rmse of 23.8 cm$^{-1}$.[41] GPR was shown to outperform a traditional NN for this system (and using the same data) in Ref. [57]. The NN with fixed, rule-based ***y*** (neuron inputs) and HDMR-GPR based neuron activation functions has a test rmse of about 20-24 cm$^{-1}$ (train set rmse of about 2.5-3.3 cm$^{-1}$), i.e., similar to a full-dimensional GPR, when the number of terms exceeds 500. This is a relatively high number of neurons, but in our method, there is no non-linear optimization – the method is as robust as a linear regression. Similar to the 3D case, once the plateau in the test rmse is reached, a larger number of terms does not lead to overfitting.

## 4 Conclusions

We showed that using a first-order additive (HDMR) model in redundant coordinates, one can obtain a single-hidden later neural network with optimal neuron activation functions for each neuron. The activation functions are the component functions of the 1$^{st}$ order HDMR and are conveniently obtained by an HDMR-GPR combination, whereby one uses an additive kernel to obtain an additive representation of the target function. The component functions computed by the HDMR-GPR model are linear combinations of smooth kernel functions; they therefore satisfy the conditions for universal approximator when considered as neuron activation functions.

In our approach, a simple, general rule-based definition of redundant coordinates – neuron arguments is used, avoiding non-linear optimization of parameters which is the main factor determining the cost of NN training and the danger of overfitting. We proposed a couple rules to define redundant coordinates that gave good results. The best results were obtained with a rule that generates new coordinates by multiplying the original coordinates by vectors-elements of a multidimensional Sobol sequence. Other definitions are possible. The pruning of the component functions / neurons has also been shown to be effective based on their variance, to reduce the final size of the model. The resulting method combines the advantage of robustness of a linear regression with the higher expressive power of a NN. We applied the method to the fitting of potential energy surfaces in the spectroscopically relevant region as an example of an application where high accuracy is required and where coupling between coordinates is significant and well understood from prior literature. We used a very last test set (much larger than the training set) to gauge the global quality of the regression. Without requiring any non-linear optimization, the NN with fixed, rule-based neuron inputs and HDMR-GPR based neuron activation functions outperforms a conventional NN in the high accuracy regime, where a conventional NN suffers more from overfitting, and obtains test set errors of less than 1 cm$^{-1}$ from 1000 training points with about 100 terms for $H_2O$, and obtains test set errors of about 20 cm$^{-1}$ from 1000 training points with about 500 or more terms for $H_2CO$, which is practically as accurate as the best full-dimensional GPR model.

# 5 Acknowledgements

This work was supported by JST-Mirai Program Grant Number JPMJMI22H1, Japan.

# 6 Data availability statement

The Matlab code used in the presented calculations are available from the corresponding author upon reasonable request. The data used are available in the supporting information of Ref. [40].